\documentclass{article}
\usepackage[utf8]{inputenc}
\usepackage{amsmath}
\usepackage{amsfonts}
\usepackage{amssymb}
\usepackage{changepage}
\usepackage{bm}
\usepackage[english]{babel}
\usepackage{natbib}
\usepackage{graphicx}
\usepackage{hyperref}
\usepackage[ruled,linesnumbered]{algorithm2e}
\usepackage{mathtools}
\usepackage{float}
\usepackage{threeparttable}
\usepackage{makecell}
\usepackage{listings}
\usepackage{subfloat}
\usepackage{bookmark}
\usepackage{color}
\definecolor{dkgreen}{rgb}{0,0.65,0}
\definecolor{gray}{rgb}{0.5,0.5,0.5}
\definecolor{mauve}{rgb}{0.8,0,0.5}

\UseRawInputEncoding

\setcitestyle{numbers}
\setcitestyle{square}
\SetArgSty{textnormal}
\renewcommand{\thefootnote}{\fnsymbol{footnote}}
\graphicspath{ {./} }
\usepackage[verbose=true,letterpaper]{geometry}
\AtBeginDocument{
  \newgeometry{
    textheight=9in,
    textwidth=5.5in,
    top=1in,
    headheight=12pt,
    headsep=25pt,
    footskip=30pt
  }
}
\DeclareMathOperator*{\argmin}{arg\,min}
\newcommand{\x}{\textbf{x}}
\newcommand{\vel}{\textbf{v}}
\newcommand{\z}{\textbf{z}}
\newcommand{\zt}{\textbf{z}_t}
\newcommand{\zprev}{\textbf{z}_{t-1}}
\newcommand{\xpred}{\x_\theta(\zt, t)}
\newcommand{\xoptimal}{\x\strut^{\hspace{-0.25ex}*}(\zt, t)}
\newcommand{\xstar}{\x\strut^{\hspace{-0.25ex}*}}
\newcommand{\at}{\alpha_t}
\newcommand{\bt}{\beta_t}
\newcommand{\qalpha}{\sqrt{\at}}
\newcommand{\oneminusa}{1 - \at}
\newcommand{\RD}{\mathbb{R}^D}

\newcommand{\xbar}{\bar{\x}(\zt)}

\newcommand{\xpreduncond}[1]{\x_\theta^{(#1)}(\zt)}

\newcommand{\posterior}{q(\zprev \vert \zt, \x)}
\newcommand{\prior}{p_\theta(\zprev \vert \zt)}

\newcommand{\brackets}[1]{\left[#1\right]}
\newcommand{\pars}[1]{\left(#1\right)}
\newcommand{\E}[2]{\mathbb{E}_{#1}\brackets{#2}}

\newcommand{\rmsnorm}[1]{\left\lVert#1\right\rVert}
\newcommand{\norm}[1]{\left\lVert#1\right\rVert_2^2}
\newcommand{\stdnormal}{\mathcal{N}(\textbf{0}, \textbf{I})}

\newcommand{\comment}[1]{}

\begin{document}

\title{Improving Diffusion Model Efficiency Through Patching}

\author{
  Troy Luhman\thanks{Equal Contribution} \\
  \texttt{troyluhman@gmail.com}
  \and
  Eric Luhman\footnotemark[1] \\
  \texttt{ericluhman2@gmail.com}
}
\date{\vspace{-5ex}}

\maketitle 

\renewcommand{\thefootnote}{\arabic{footnote}}


\begin{center}
\textbf{Abstract}
\\
\end{center}
\begin{changemargin}{1.75cm}{1.75cm} 
Diffusion models are a powerful class of generative models that iteratively denoise samples to produce data. While many works have focused on the number of iterations in this sampling procedure, few have focused on the cost of each iteration. We find that adding a simple ViT-style patching transformation can considerably reduce a diffusion model's sampling time and memory usage. We justify our approach both through an analysis of the diffusion model objective, and through empirical experiments on LSUN Church, ImageNet $256^2$, and FFHQ $1024^2$. We provide implementations in Tensorflow\footnote{https://github.com/ericl122333/PatchDiffusion-TF} and Pytorch\footnote{https://github.com/ericl122333/PatchDiffusion-Pytorch}.
\end{changemargin}


\section{Introduction}
Since their introduction in \cite{originaldiffusion}, diffusion models have gained increased attention due to their high-fidelity samples, good data coverage, and straightforward implementation \cite{ddpm}. On image data, diffusion models were shown to surpass state-of-the-art GANs \cite{originalgan, stylegan, biggan} in visual quality \cite{adm}, and have achieved unprecedented performance on text-to-image tasks \cite{dalle2, imagen}. Futhermore, pretrained diffusion models are well-suited for downstream tasks such as image inpainting \cite{repaint} or stroke-to-image synthesis \cite{sdedit}, often outperforming rival methods without any task-specific training.

However, this success comes at the cost of a high computational load. At sampling time, they usually require hundreds-to-thousands of network evaluations to simulate the reverse diffusion process. In addition, their mode-covering likelihood objective requires a large amount of training compute for complex, high-resolution datasets like ImageNet \cite{imagenet}. For instance, training an ADM \cite{adm} on 256 $\times$ 256 ImageNet uses 914 V-100 days of compute, which can be costly and inaccessible.

For image data, an effective way of reducing compute costs is to model a low-dimensional representation of the data, which can be obtained through downsampling \cite{iddpm, adm, cdm} or a latent encoder \cite{vqgan, ldm}. However, the compressed data must be decoded using a second network, resulting in a complicated pipeline that changes the original diffusion process. This modification pushes an extra layer of considerations onto practitioners looking to apply diffusion models for downstream tasks \cite{ilvr, palette, styleclip}.

In this work, we propose to apply diffusion models on a grid of image patches, where each patch represents a group of neighboring pixels. This ``patching" operation moves pixels from the spatial dimension to the channel axis, leading to significant improvements in both efficiency and memory usage. Furthermore, patching is simple to implement and leaves the diffusion processes statistically unchanged, making it easily applicable to any diffusion model. 

Our key contributions can be summarized as follows:
\begin{itemize}
\item{We present an in-depth analysis of the diffusion model objective, and show that high-resolution convolutional layers are redundant at most timesteps.}
\item{We present a new patched diffusion model (PDM), and show that it greatly reduces the compute requirements for generating high-resolution images. Additionally, it's a sampling speed movement orthogonal to existing methods, as it concerns the function itself, not the number of function evaluations.}
\item{We study the effect of the output parameterization in diffusion models, both with and without patching, and find that data prediction is less fragile than noise prediction.}
\item{We propose a method for scaling PDMs to difficult datasets by splitting a single network into multiple parts; and demonstrate high-fidelity image synthesis on the challenging ImageNet $256^2$ dataset.}
\end{itemize}

\section{Patched Diffusion Models}

\subsection{Diffusion Models}
\label{section2.1}
We provide a brief review of the ``variance-preserving" diffusion models from \cite{ddpm}. Consider a forward process $q$, indexed by timestep $t$, that produces latents $\z_1 \ldots \z_T$ given data $\x \in \RD$. The forward transition kernels can be written as follows:
\begin{align}
q(\zt \vert \zprev) \coloneqq & \ \mathcal{N} \pars{\sqrt{1-\bt}\zprev, \bt \textbf{I}}, \ \bt \in (0, 1) \label{eq:1}\\
q(\zt \vert \x) = & \ \mathcal{N} \pars{\qalpha\x, (\oneminusa) \textbf{I}}, \ \at \coloneqq \prod_{s=1}^{t} (1-\beta_s) \label{eq:2}
\end{align}
With this, we can compare the posterior $\posterior$ to a learned prior $\prior$ for maximum likelihood training, in a similar fashion to \cite{vae}. Since $\posterior$ is Gaussian for all $t$, we set $\prior$ to be Gaussian with mean $\mu_\theta(\zt, t)$ and the same variance as $\posterior$.\footnote{Learned variances are also possible; see Appendix \ref{implementationdetails} for details.} This allows for closed form comparison of KL divergences between the two. Sampling from the generative model is straightforward, starting with a Markov chain initialized at $p(\z_T) = \stdnormal$, and continuing with each $\prior$ in succession.

To train $\mu_\theta(\zt, t)$, one could simply predict the mean of $\posterior$ directly. However, the mean of $\posterior$ becomes very close to $\zt$ itself for large $T$, decreasing the model's learning signal. As such, \cite{ddpm} found it beneficial to parameterize $\mu_\theta(\zt, t)$ in terms of $\epsilon_\theta(\zt, t)$, which predicts the noise $\epsilon_t = (\zt - \qalpha \x) / \sqrt{\oneminusa}$. We instead choose to predict the data $\x$ directly, and we further discuss this choice in Section \ref{section3.2}.

An evidence lower bound on $\log q(\x)$ can now be written as a weighted sum of denoising autoencoder objectives, where the network $\xpred$ learns to reconstruct data at multiple levels of noise:
\begin{equation}\label{eq:3}
-\text{ELBO} =  \sum_{t=1}^{T} \gamma_t \E{q(\x, \zt)}{\norm{\xpred - \x}} + C
\end{equation} 
where $\gamma_t$ is a constant related to $\bt$, $\at$, and $\alpha_{t-1}$. In practice, optimizing  Equation \ref{eq:3} with the maximum-likelihood weighting of $\gamma_t$ typically leads to lower sample quality, as it overemphasizes high-frequency details \cite{ddpm}. In our experiments, we set $\gamma_t = \sqrt{\frac{\at}{\oneminusa}}$ to prioritize learning of low-level features.

\subsection{Redundancies in Existing Model Architectures}
\label{subsec2.2}
Optimization of Equation \ref{eq:3} is typically done by sampling $\x$ from the dataset, perturbing it into noisy image $\zt$, then training the model to predict the original $\x$. However, a perfect reconstruction is an impossible task, since $\zt$ \textit{could} have originated from a different value of $\x$. Indeed, any example in the dataset has the potential produce a given $\zt$, just not with equal probability. As such, the best strategy for $\xpred$ would be to predict a weighted average of dataset examples, where each is weighted according to the probability that it generated $\zt$.

Expressed mathematically, the optimal value of $\xpred$ that minimizes Equation \ref{eq:3} can be written as: \begin{equation}
\label{eq:4}
\xoptimal = \E{q(\x)}{\x \ \frac{q(\zt \vert \x)}{q(\zt)}}
\end{equation} which also equals the mean of $q(\x \vert \zt)$. See derivation in Appendix \ref{appendix:A}. Despite being the reconstruction target that minimizes Equation \ref{eq:3}, values of $\xoptimal$ actually lie on a lower-dimensional manifold than the data itself. Instead of resembling actual images, $\xoptimal$ looks like a blurred version, where the blurring increases with time  (Figure \ref{figure1}).

This effect arises from the averaging of different $\x$ in the expectation from Equation \ref{eq:4}. Because the weighting factor $q(\zt \vert \x)$ assigns higher probability to $\x$ that are close to $\zt$ in pixel space, the most likely values of $\x$ will have similar low-frequency components. However, these values will have dissimilar high-frequency components; these high-frequency components then destructively interfere with each other, causing blur (see Figure \ref{figureb}). As $t$ increases, values of $\x$ that are far away from $\zt$ have a higher weighting in the expectation, increasing the blurriness of $\x^{\text{*}}$\hspace{-0.5ex}.

Thus, when choosing neural network architectures for $\xpred$, it may be helpful to consider which layers are most helpful in reaching the desired minimum at $\xoptimal$. Typically, diffusion models use convolutional layers at multiple resolutions to model both low and high-frequency components. But since the high-frequency components in $\xoptimal$ are attenuated early in the forward process, we argue that high-frequency convolutional layers are redundant for most timesteps. Additionally, these layers often have the largest memory consumption because they operate at high spatial resolutions.

\begin{subfigures}
\begin{figure}
\vspace*{-0.3cm}
  \includegraphics[scale=0.32]{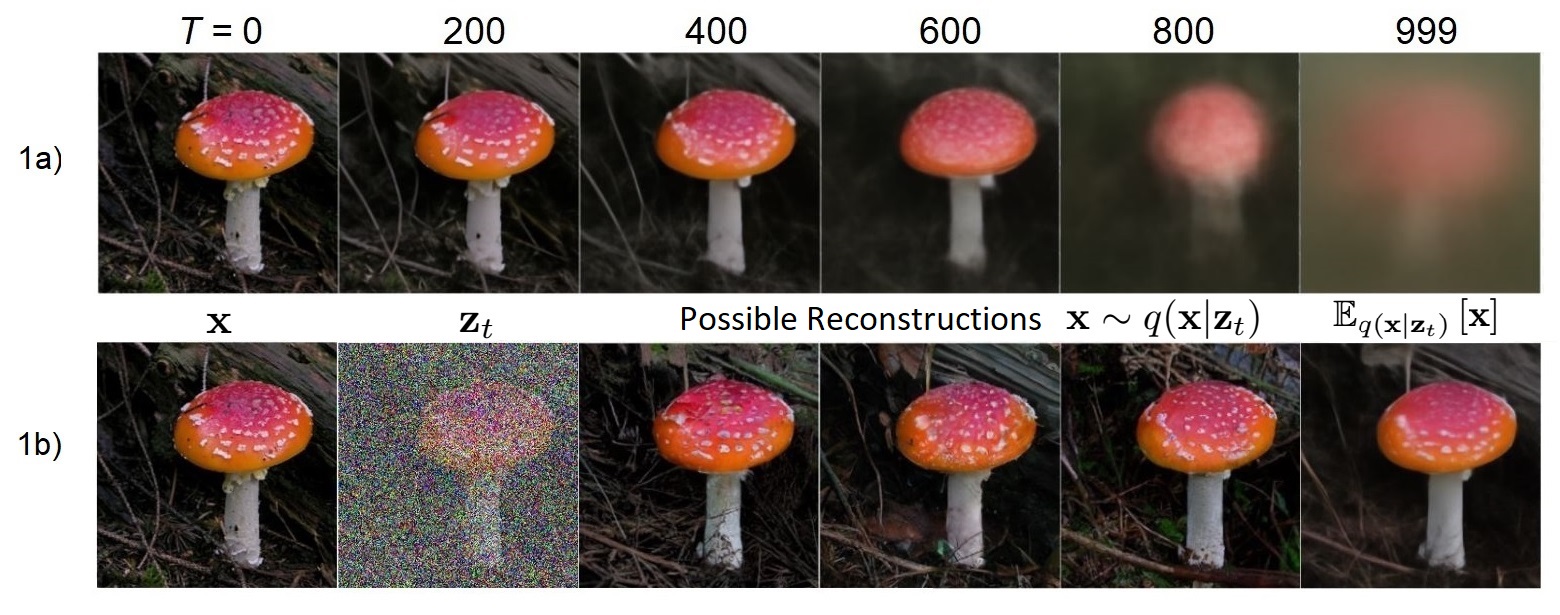}
  \caption{\label{figure1} \hspace*{-0.2cm} The value of $\xpred \approx \xoptimal$ as $t$ increases.
  }
\end{figure}
\begin{figure}
\vspace*{-0.5cm}
\caption{\label{figureb} Possible reconstructions sampled from noisy image $\zt$. We approximate $q(\x \vert \zt)$ using a trained ImageNet model.}
\end{figure}
\end{subfigures}

\section{Patched Diffusion Models}


The standard U-Net architecture for diffusion models takes in inputs $\zt$ with shape ($H$, $W$, $C$), where $H$, $W$ and $C$ are the height, width and number of channels of the image. Motivated by the discussion in Section \ref{subsec2.2}, we propose to reshape the image into a grid of non-overlapping patches with shape ($H/P$, $W/P$, $C \times P^2$) where $P$ is the patch size. This eliminates potentially wasteful computation at the high noise levels, and within-patch dependencies can still be modeled via channel-wise operations.

\subsection{Effect of patch size}
An important consideration in using patching is how large the patch size $P$ should be. Learning diffusion models on patches of image data enables a significant reduction in computational resources. However, the number of output channels grows quadratically with $P$, and it might be difficult to predict all the pixels in a $P \times P \times C$ grid without significantly increasing the network's width. 

To test how much we can increase the patch size without significantly reducing quality, we train 3 different models on the LSUN Churches $256 \times 256$ dataset \cite{lsun}, with patch sizes $P = 2, 4, \text{and } 8$. Each model has roughly 120 million parameters and is trained for 2.5 days on a TPUv2-8 (approx. 5 V100 days). 

In Table \ref{tab:table1}, we report both the sample quality measured by FID \cite{fid} and the sampling speed. We find that $P = 4$ achieves a good trade-off between efficiency and quality, so we use this choice for our other models. We show a visual comparison for the different models in Appendix \ref{appendixE}.

We also investigate how different choices of $P$ affect the distortion  $\rmsnorm{ \x - \xpred}$ at various timesteps. As shown in Figure \ref{fig2distortion}, the distortion is noticeably higher for larger patch sizes when $t$ is low. However, for the majority of timestep values, models with $P = 4, 8$ have similar or even lower levels of distortion compared to the more computationally expensive $P = 2$ model. This result is consistent with our analysis in \ref{subsec2.2}, as it demonstrates that high-resolution convolutions are not necesary when $\xoptimal$ itself is blurry.

\begin{minipage}{0.42\linewidth} 
\begin{table}[H]
  \begin{center}
    \caption{Effect of varying the patch size on FID and sampling speed. Sampling speed is evaluated on a single TPUv2-8 with a batch size of 1024 and 250 sampling steps.} 
    \label{tab:table1}
    \begin{tabular}{c c c} 
      \vspace*{-0.15cm} \\ 
      Patch Size &\hspace*{-0.25cm} FID $\downarrow$ &\hspace*{-0.2cm} Images/sec $\uparrow$ \\
      \hline \vspace*{-0.15cm} \\ 
      \vspace{0.15cm} $P = 2$ & 5.7 & 1.7 \\ \vspace{0.2cm}
      $P = 4$ & 7.5 & 2.4 \\ \vspace{0.2cm}
      $P = 8$ & 22.0 & 4.4 \\ \vspace{0.15cm}
      
    \end{tabular}
  \end{center}
 
\end{table}
\end{minipage}
\hspace{0.5cm}
\begin{minipage}{0.5\linewidth} 
\begin{figure}[H]
\includegraphics[scale=0.5]{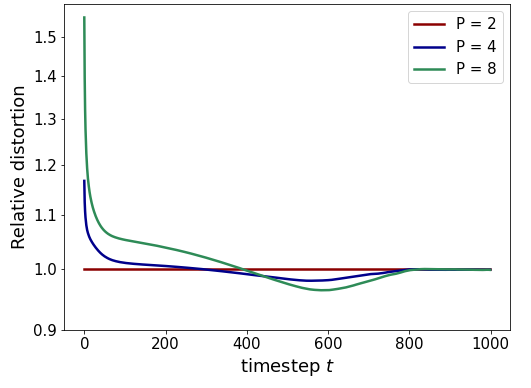}
\vspace*{-0.3cm}
\caption{Ratio of distortion (RMSE) when using $P = 4, 8$ compared to $P = 2$}
\label{fig2distortion}
\end{figure}
\end{minipage}

\subsection{Output parameterization}
\label{section3.2}
From its definition in section \ref{section2.1}, the reconstruction network $\xpred$ was trained to predict the data $\x$ itself, instead of the noise $\epsilon$ that was added to it. We hypothesize that patch-predicting models will benefit from this approach, since pixels within a patch of $\x$ will generally have similar values, while the same is not true for $\epsilon$. To test this hypothesis, we report initial loss curves for models that output $\x$ and $\epsilon$, as well as $\vel$ \cite{progdist} where $\vel \coloneqq \sqrt{\at}\epsilon - \sqrt{1-\at}\x$. We include results for both patched and un-patched models in Figure \ref{figure3}.

Interestingly, while patched models indeed perform better with $\x$-prediction compared to $\epsilon$-prediction, $\vel$-prediction performs similarly well despite the larger within-patch variance. We hypothesize this is due to the learning signal the model receives. With $\epsilon$-prediction, the model must learn a near-identity function at high timesteps; the same is true with $\x$-prediction but at the low timesteps. However, at high timesteps, small errors in the prediction of $\epsilon$ induce large errors in the predicted $\x$ because of division by $\sqrt{\alpha}$; see figure \ref{figure4} for an example. Therefore, we recommend that patched models use $\x$ or $\vel$ prediction, and believe such parameterization may be beneficial in other settings as well.

\begin{minipage}{0.48\linewidth} 
\begin{figure}[H]
\includegraphics[scale=0.33]{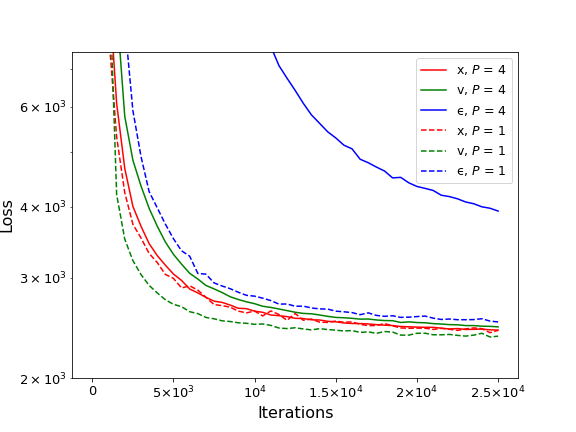}
\vspace{-0.6cm}
\caption{Loss when using different prediction types, both for $P = 4$ and $P = 1$. }
\label{figure3}
\end{figure}
\end{minipage}
\hspace{0.25cm}
\begin{minipage}{0.48\linewidth}
\begin{figure}[H]
\includegraphics[scale=0.17]{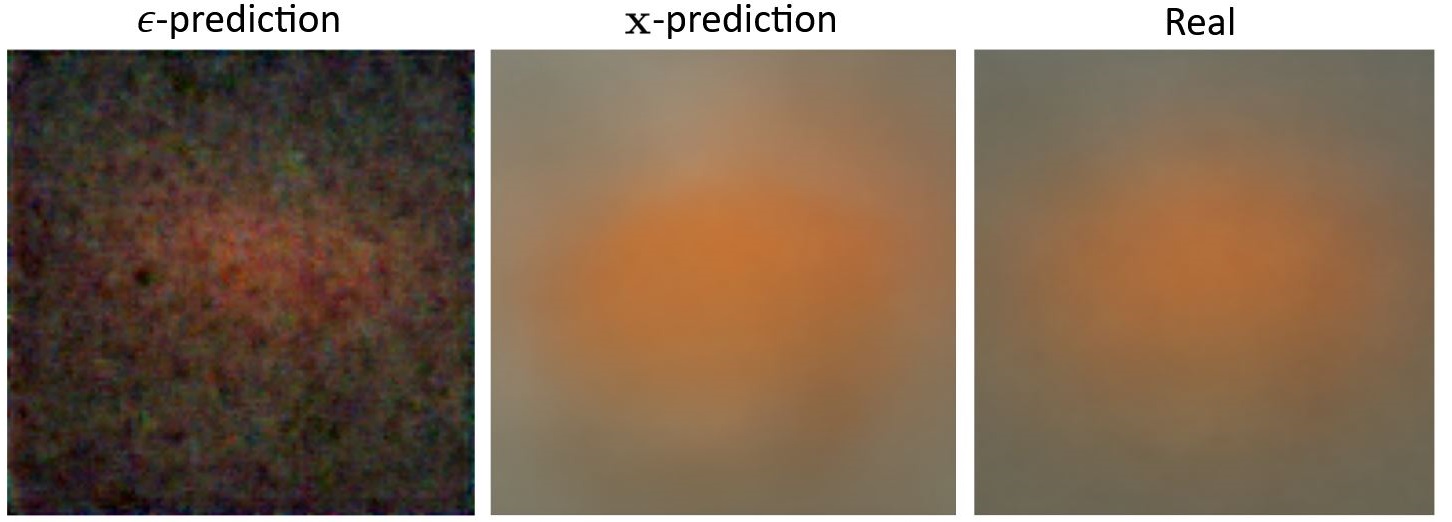}
\caption{Estimates for $\x_\theta$ from $\epsilon$-predicting (ADM) and $\x$-predicting (PDM) networks, and the true $\xstar$, at $t=970$. $\epsilon$-prediction produces noise artifacts in $\x_\theta$. We approximated $\xstar$ using the empirical distribution for the class ``goldfish".}
\label{figure4}
\end{figure}
\end{minipage} 

\section{Scaling to more complex datasets}
In this section, we evaluate the performance of Patched Diffusion Models on ImageNet $256 \times 256$, and on the FFHQ $1024 \times 1024$ dataset. Both experiments use $P = 4$.

\begin{table}[H]
  \begin{center}
    
    \label{tab:table2}
    \begin{tabular}{l c c c c c c} 
      \\
      Method&\hspace{-0.15cm}Training Time&IS $\uparrow$&FID $\downarrow$&sFID $\downarrow$& Precision $\uparrow$&Recall $\uparrow$ \\
      \hline \\
      PDM (Ours), $w = 1.5$ & 32 & 142.2 & 8.57 & 6.50 & 0.79 & 0.47 \\
      PDM (Ours), $w = 2.25$ & 32 & 244.7 & 8.87 & 6.84 & \textbf{0.90} & 0.31 \\
      
      ADM-G, $w = 1.0^{\dagger}$ & 962 & 186.7 & \textbf{4.59} & \textbf{5.25} & 0.82 & 0.52 \\
      ADM-G, $w = 10.0^{\dagger}$ & 962 & \textbf{283.92} & 9.11 & 10.93 & 0.88 & 0.32 \\
      ADM-G-U, $w = 1.0^{\dagger}$ & 349 & 215.84 & 3.94 & 6.14 & 0.83 & \textbf{0.53} \\
      
      LDM-8-G, $w = 1.0^{\dagger}$ & 91 & 190.4 & 8.11 & - & 0.83 & 0.36 \\
      LDM-4-G, $w = 1.5^{\dagger}$ & 271 & 247.67 & 3.60 & - & 0.87 & 0.48 \\
      
      BigGAN & 128-256 & 202.65 & 6.95 & 7.36 & 0.87 & 0.28 \\
      StyleGAN-XL & - & 265.1 & 2.30 & 4.02 & 0.78 & \textbf{0.53} \\
      
    \end{tabular}
     \caption{
    ImageNet $256 \times 256$ results and compute comparison. All values for comparison were taken from their respective works; unreported values are indicated with a ``-". Training time is measured in V100-days; we use the conversion 1 TPUv2-8 day = 2 V100-days. \newline$\dagger$ indicates classifier guidance. All diffusion models use 250 sampling steps except LDM-8-G which uses 100. }
    
  \end{center}

\end{table}

\subsection{Model Splitting}
For high-resolution, multimodal datasets, previous works have benefited from using two specialized networks for a single task. One creates a downsampled version of the data while the other upsamples these images \cite{iddpm, adm, cdm}. As our models already operates on inputs with smaller height and width, we propose a different approach for two-stage pipelines. Specifically, our ``model splitting" method uses two different models on the full-dimensional data, where the first learns to denoise $\zt$ at timesteps 1 to $S$, while the second denoises for timesteps $S \text{ to } T$. 

Similar to upsampling pipelines, model splitting implicitly divides the generative process into creating a ``low-resolution" image, and a high-resolution image from this. In our case, the ``low-resolution" image is created through the blurring effect intrinsic to the forward process (as described in Section \ref{subsec2.2}) and preserves the dimensionality of the data. To analyze these contrasting approaches through a different lens, a downsampling kernel averages over regions of the same image, while the forward process averages images with other samples from the dataset.

Compared to other two-stage approaches, partioning the problem according to the forward process is advantageous because of its flexibility and simplicity. By avoiding external latent variables, it retains easy likelihood computation and can be applied to downstream tasks in the exact same manner as an un-split model. 

\subsection{ImageNet Results}

To test the scalability of our approach, we conduct an experiment on the challenging ImageNet \cite{imagenet} dataset at the $256 \times 256$ resolution. We use model splitting with $S = 396$ (SNR $\approx 0.25$). Both models use identitical architectures, allowing us to initialize the second model's weights with the first model's weights to warm-start the training process. Our sampling procedure uses dynamically thresholded classifier-free guidance \cite{classifierfree, imagen} at weights of $w = 1.5$ and $w = 2.25$, and 250 sampling steps (500 function evaluations). 

Table \ref{tab:table2} compares the performance and training costs of our approach to various other DDPM and GAN based approaches. Each split is trained for 500 thousand iterations on a batch size of 256. On a single TPUv2-8, we achieve a throughput of 183 images per second, or 1.4 seconds per iteration. Despite using vastly fewer computational resources, our model achieves competitive FID \cite{fid} with other methods. Especially surprising is our low spatial FID (sFID) \cite{sfid}; our model performs relatively better at modeling high-level relationships compared to large-scale structure, even without convolutional layers at the $256^2 \text{ and } 128^2$ resolutions. It is likely that all these metrics would improve with longer training.

\subsection{FFHQ}

To demonstrate that patching in diffusion models is effective at higher resolutions, we evaluate the quality of generated $1024^2$ faces on FFHQ. As shown in Figure 5, when using $P = 4$, the model is able to create images with sharp pixel-level details. 

An important consideration when working with higher resolutions such as $1024^2$ is memory usage, so we compare our FFHQ model's memory footprint to a baseline that uses $P = 1$ \cite{scoresde}. On a single P100 using single precision, the maximum batch size our model could fit without running out of memory was 32, while the baseline with $P = 1$ could only hold a batch size of 8. This corresponds to a $4 \times$ decrease in memory cost, which is reasonable considering our model takes in inputs with a $4 \times$ smaller height and width. By increasing the maximum batch size, PDMs gain an additional speedup on batch inference tasks over a model without patching.

\begin{figure}[H]

\begin{measuredfigure}
\includegraphics[scale=0.45]{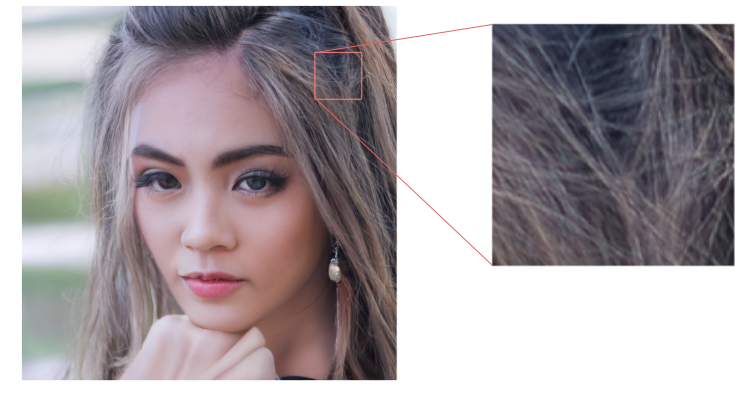}
\vspace*{-0.3cm}
\caption{A generated sample from our FFHQ model with $P = 4$. Even when zooming in, there are no discernable artifacts that may have resulted from using patching.}
\label{ffhqzoomin}
 \end{measuredfigure}
 
\label{fig2ffhqzoomin}
\end{figure}

\section{Related Work}

Our work is most closely related to Subspace Diffusion models \cite{subspaceddpm}, who add dimensionality-reducing projections at specified points of diffusion process. Their framework allows for improved efficiency without marginalizing over external latent variables, but it nevertheless modifies the diffusion process and requires using multiple models. In contrast, our method acts on the full dimensional data and achieves speedup through an architecture change. Other similar methods, such as cascading \cite{cdm} or latent diffusion \cite{ldm}, model lower-dimensional data and use a multi-stage generation process to produce full-resolution images. 

For generative models, a variant of patching was introduced in the coupling layers of normalizing flows \cite{realnvp} to partition the data dimensions along the spatial axes, enabling a computationally tractable determinant. Patching also sees widespread usage in vision transformers \cite{vit}, which have seen success on various other image tasks \cite{objectdetect, segmenter}. ViT-based architectures have also been applied to GANs \cite{vitgan, vitvqgan}, but not to diffusion models; we briefly experimented with ViTs but found that U-Nets generally outperformed them (details in Appendix \ref{vitdetails}).

\section{Conclusion}
Our work provides a practical solution to reduce training and sampling costs through the use of patching. Unlike previous works that achieve efficiency by using compressed data representations, our method works directly on the original data, and is applicable to any diffusion model. By adjusting the patch size, practitioners can trade off speed for quality depending on their resource constraints. Alternatively, one can avoid this trade-off by using multiple models with different patch sizes for different noise levels; combining such an approach into a single model is a possible direction of future work. Another avenue for future work is applying vision transformers to diffusion models, since transformer-based architectures are often more powerful and scalable than CNN-based architectures.

\pagebreak

\bibliography{references}

\pagebreak

\appendix

\section{Derivations}
\label{appendix:A}
This appendix contains the derivation for Equation \ref{eq:4}, the formula for the optimal reconstruction network that minimizes Equation \ref{eq:3}. First, define mapping function $\x_\theta : \RD \times \{1, \ldots, T\}  \rightarrow  \RD$, whose cost function $J(\theta)$ has its global minimum at $\x^{\text{*}}.$ 
\begin{align}
J(\theta) \coloneqq \sum_{t=1}^{T} & \gamma_t \E{q(\x, \zt)}{\norm{\xpred - \x}} + C \\
\x^{\text{*}} & \coloneqq \argmin_\theta J(\theta)
\end{align}
We now show that for any $\zt \in \RD$ and $t \in \{1, \ldots, T\}$, we have $\xoptimal = \E{q(\x)}{\x \ \frac{q(\zt \vert \x)}{q(\zt)}}$. Because the function $\xpred$ receives timestep $t$ as an input, we can treat $\xpred$ as \newline $T$ different functions $\xpreduncond{1}, \ \ldots \ \xpreduncond{T}$, and have each $\xpreduncond{t}$ optimize a single term of the summation. We now turn our attention to finding the minimization of each individual term. Defining $\xbar \coloneqq \E{q(\x \vert \zt)}{\x}$, we have: 
\begin{align}
{\x^{(t)}}^{\text{*}} & = \argmin_\theta \gamma_t \E{q(\x, \zt)}{\norm{\xpreduncond{t} - \x}} \\
& = \argmin_\theta \gamma_t \E{q(\zt)}{ \E{q(\x \vert \zt)}{\norm{\xpreduncond{t} - \x}} } \\
& = \argmin_\theta \gamma_t \E{q(\zt)}{ \E{q(\x \vert \zt)}{\norm{\pars{\xpreduncond{t} - \xbar} - \pars{\x - \xbar}}} } \\
& = \argmin_\theta \gamma_t \E{q(\zt)}{ \E{q(\x \vert \zt)}{\norm{\xpreduncond{t} - \xbar} + \norm{\x - \xbar} }} \\
& = \argmin_\theta \gamma_t \E{q(\zt)}{ \E{q(\x \vert \zt)}{\norm{\xpreduncond{t} - \xbar} }} \\
& = \argmin_\theta \gamma_t \E{q(\zt)}{\norm{\xpreduncond{t} - \xbar}}
\end{align}
From this, we can see that ${\x^{(t)}}^{\text{*}}$ will always map $\zt$ to $\xbar$, or the mean of $q(\x \vert \zt)$. This makes sense, since the mean-squared loss over $q(\x \vert \zt)$ in Equation 8 is minimized by its mean. Finally, the result in Equation \ref{eq:4} can be obtained by using Bayes' rule. 

Given the connections between diffusion models and score-based models \cite{ncsn, ddpm, scoresde}, we offer an alternative interpretation of our result in Equation \ref{eq:4} through the lens of score-matching \cite{originalsm}. Specifically, when setting $\nabla_{\zt} \log q(\zt) = -(\zt - \sqrt{\at}\xbar)/(1-\at)$ the objective in Equation 12 becomes a score-matching objective between score network $s_\theta^{(t)}(\zt)$ and the marginal score. Meanwhile, Equation 7 corresponds to the denoising score matching objective which is equivalent to the score matching objective up to a constant \cite{vincent2011}.

\pagebreak
\section{Code Implementation}
In this appendix, we provide an example PyTorch implementation of our patched architecture, that includes code segments for both the patching and unpatching operations. Aside from the architecture changes listed here, no other modifications are necessary. Note that other implementations of patching may exist; this one uses the same patch ordering as Tensorflow's \texttt{image.extract\_patches} function.

\begin{lstlisting}
# Code for a modified UNet architecture that operates on patched images
class PatchedUNet(nn.Module):
	def __init__(self, *args, **kwargs):
		# create the U-Net architecture as normal 
		# but make sure input/output projections use 3*patch_size**2 channels, not 3
		# ...
		self.patch_size = patch_size
   	    
	def convert_image_to_patches(self, x):
		p = self.patch_size
		B, C, H, W = x.shape
	
		x = x.permute(0, 2, 3, 1)       #BHWC format, bc reshape is done on last 2 axes
		x = x.reshape(B, H, W//p, C*p)  #reshape from width axis to channel axis

		x = x.permute(0, 2, 1, 3)           #now height & channel should be last 2 axes
		x = x.reshape(B, W//p, H//p, C*p*p) #reshape from height axis to channel axis
		return x.permute(0, 3, 2, 1)        #convert to channels-first format
        
	def convert_patches_to_image(self, x):
		p = self.patch_size
		B, C, H, W = x.shape

		x = x.permute(0,3,2,1) #BWHC; from_patches starts w/ height axis, not width
		x.reshape(B, W, H*p, C//p)     #reshape from channel axis to height axis

		x = x.permute(0,2,1,3)  #now width & channel should be last 2 axes
		x = x.reshape(B, H*p, W*p, C//(p*p)) #reshape from channel axis to width axis
		return x.permute(0, 3, 1, 2)         #convert to channels-first format
		
	def forward(self, x, t):
        x = self.convert_image_to_patches(x)

        # run the forward pass of the U-Net the same exact way as before.
        # ...

        x = self.convert_patches_to_image(x)
        return x
    
\end{lstlisting}

\pagebreak
\section{Implementation Details} \label{implementationdetails}

This section covers some of the most significant choices we made during experimentation. For more detailed information, one can visit our Tensorflow implementation or a Pytorch implementation which is based on the repository from \cite{adm}. Our U-Net architecture is similar to the one used in \cite{adm}, but with several changes that we list below:

\begin{itemize} 
\item{Similar to \cite{iddpm}, we learn the reverse process variances to facilitate faster sampling. However, instead of optimizing their hybrid objective, we chose to use the main network to optimize $L_\text{simple}$ only, and use a small auxiliary network to minimize $L_\text{vlb}$. We opted for this approach to avoid having to balance out the two different objectives, as $\lambda=0.001$ may not be suitable for our different weighting of $\gamma_t$ in $L_\text{simple}$.}
\item{We add 2D sinusoidal positional encodings \cite{transformer, vit} to incorporate spatial information into the network. This information is added once, right after the first input projection.}
\item{We eliminate attention at the $32^2$ resolution due to memory constraints. Before the $16^2$ and $8^2$ attention, we also add a GroupNormalization \cite{groupnorm} layer to normalize attention inputs.}
\item{In the upsampling stack, we decrease the kernel size of the convolutional skip connection from 3 to 1 for reduced parameter counts.}
\item{For class-conditional experiments on ImageNet, we injected class information through a sigmoid gate before the residual connection in our convolutional blocks, instead of in the AdaGN layers.}
\item{To reduce memory requirements in distributed training, we held a single copy of the exponential-moving-averaged weights on the CPU instead of replicating across devices. For better throughput, we updated the EMA weights once every 100 iterations, with decay of 0.99.}
\end{itemize}

\begin{table}[h!]
  \begin{center}
    \caption{Hyperparameters used in our LSUN experiment.} 
    \label{tab:hparams}
    \begin{tabular}{l c c c} 
      \\ 
      Patch Size & $P = 2$ & $P = 4$ & $P = 8$\\
      \hline \\
      Parameters & 123M & 121M & 121M \\
      Base channel size & 128 & 256 & 256\\
      Channel multiplier & 1,2,2,4,4 & 1,1,2,2 & 1,1.5,2 \\
      Blocks per resolution & 2 & 2 & 3\\
      Attention resolutions & 16,8 & 16,8 & 16,8 \\
      Dropout & 0.0 & 0.0 & 0.0 \\
      Batch size & 128 & 256 & 512 \\
      Learning Rate & $1 \times 10^{-4}$ & $1 \times 10^{-4}$ & $1 \times 10^{-4}$ \\
      Adam $\beta_1$ & 0.9 & 0.9 & 0.9 \\
      Adam $\beta_2$ & 0.99 & 0.99 & 0.99 \\
      Warmup steps & 5000 & 5000 & 5000 \\
      Training iterations & 335k & 275k & 215k \\
      Diffusion $\beta_T$ & 0.02 & 0.02 & 0.02
      
    \end{tabular}
  \end{center}
\end{table}

\begin{table}[h!]
  \begin{center}
    \caption{Hyperparameters used in our experiments.} 
    \label{tab:hparams}
    \begin{tabular}{l c c c c} 
      \\ 
      Dataset & FFHQ & ImageNet $256^2$ ($1^{\text{st}}$ split) & ImageNet $256^2$ ($2^{\text{nd}}$ split)\\
      \hline \\
      Parameters & 163M & 202M & 202M \\
      Base channel size & 128 & 256 & 256\\
      Channel multiplier & 1,2,2,4,4,4 & 1,2,2,2 & 1,2,2,2 \\
      Blocks per resolution & 2 & 3 & 3\\
      Attention resolutions & 16,8 & 16,8 & 16,8 \\
      Dropout & 0.0 & 0.0 & 0.0 \\
      Batch size & 32 & 256 & 256 \\
      Learning Rate & $2 \times 10^{-5}$ & $1 \times 10^{-4}$ & $5 \times 10^{-5}$ \\
      Adam $\beta_1$ & 0.9 & 0.9 & 0.9 \\
      Adam $\beta_2$ & 0.98 & 0.98 & 0.999 \\
      Warmup steps & 5000 & 5000 & 10000 \\
      Training iterations & 780k & 500k & 500k \\
      Diffusion $\beta_T$ & 0.025 & 0.02 & 0.02
      
    \end{tabular}
  \end{center}
\end{table}

\pagebreak
\section{Vision Transfomer Experiments} \label{vitdetails} 
Arguably the most common usage of patching is in Vision Transformer architectures \cite{vit}. We tested applying these architectures to diffusion models as well, but unfortunately it was unable to achieve the same level of performance as the U-Net \cite{unet} architecture. The primary cause of this was unstable training; most models trained stably up to a point, after which their gradient diverged (to infinity in some cases). In general, divergence was only temporarily delayed by various stability tricks we tried. In the interest of completeness and potential future research, we describe our results here.

Our initial configuration trained on LSUN Church was an 11-layer ViT with $32^2$ sequences, a patch size of 8, and a batch size of 128. It used time-conditional layer normalization \textit{after} (i.e. PostNorm) each residual connection, and had linear projections of its input injected to it in at the 4th, 8th, and last layers (to mimic the long-range skips in the U-Net). However, the model's gradient norm diverged to infinity after only a few thousand iterations, even with a low value of $\beta_2 = 0.95$ and a warmup of 10000 steps. We then employed the following stability tricks:
\begin{itemize}
\item{We moved the time-conditional LayerNorm to before the attention and feedforward networks (i.e. PreNorm), and employed an initial weight scaling of $1/\sqrt{N_{layers}}$ compared to the default initializer. This configuration also diverged quickly, but less so compared to the previous.}
\item{Instead of scaling the initial weights of the output projection, we scaled the output of the residual layer itself by $1/\sqrt{N_{layers}}$, which reduces the effective learning rate of the output projection. We also switched from LayerNorm to GroupNorm. Both improved stability but could not stop eventual divergence.}
\item{To increase the model's learning signal, we switched from $\x$ prediction to $\vel$ prediction. To give some sense of convolutional information, we gave the model additional inputs that were blurred versions of its input (blurred with $\sigma=1, 2, 4, \text{and } 8$). Again, these improved stability but still resulted in divergence.}
\item{We reduced the learning rate from $1 \times 10^{-4}$ to $5 \times 10^{-5}$, and increased the warmup steps to $20000$. This configuration did not diverge, but had significant loss spikes that couldn't fully recover. We tried reducing it further to $2 \times 10^{-5}$ which trained stably, but had unacceptably high loss. }
\item{We tried using time-conditional GroupNorm in the feed-forward only, using unconditional GroupNorm before the attention. We also tried using unconditional GroupNorm in both, injected timestep information once in the beginning. The former still resulted in loss spikes; the latter did not, but had very poor performace.}
\end{itemize}

In some instances, ViT models performed comparably to a U-Net with $P = 8$ in the stage before instabilities occured. We believe that addressing the underlying instabilities can enable ViTs to reach similar or better performance compared to U-Nets.

\pagebreak
\section{Samples}
\label{appendixE}

\begin{figure}[H]
\begin{center}
\begin{measuredfigure}
\centering
\includegraphics[scale=0.18]{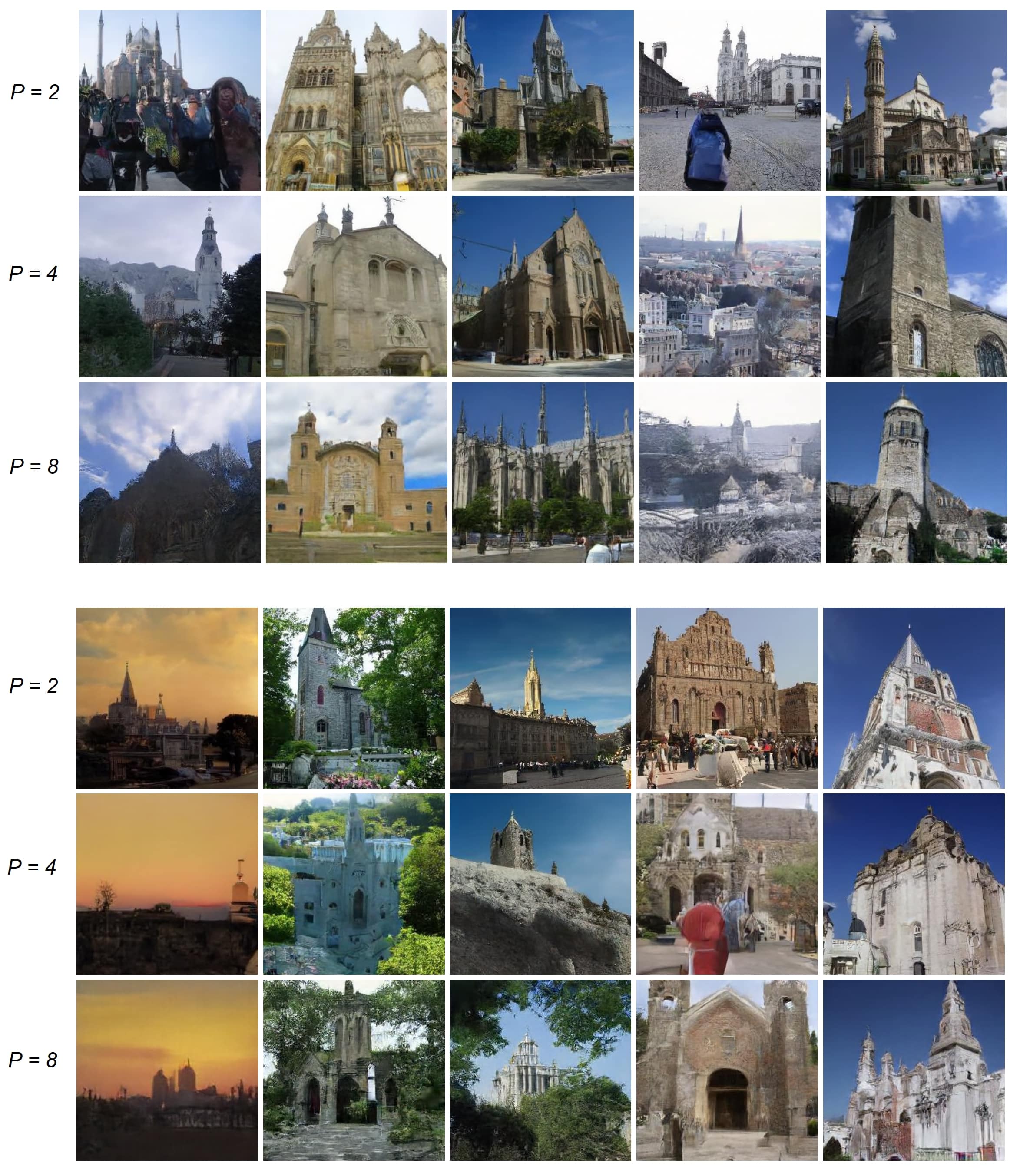}
\vspace*{-0.3cm}
\caption{A comparison of samples generated from our LSUN models with $P = 2, 4, \text{and } 8$. While images have proper global structure, images from the $P = 8$ model are sometimes blurry. We keep the noise seed constant and use the strided sampling schedule from \cite{adm}.}
\label{lsunfig}
\end{measuredfigure}
\end{center}
\end{figure}

\begin{figure}
\begin{center}
\begin{measuredfigure}
\includegraphics[scale=0.3]{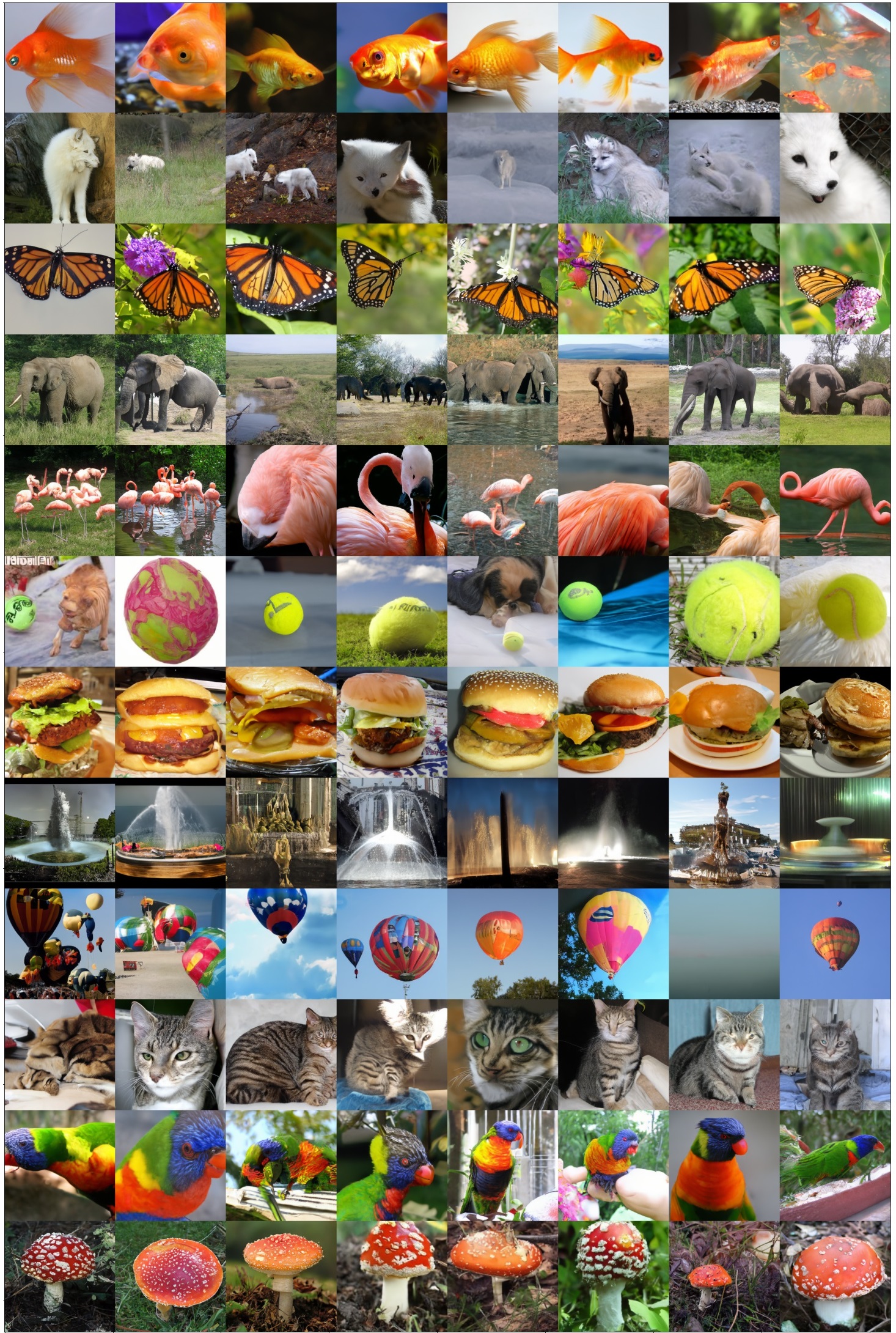}
\vspace*{-0.3cm}
\caption{Uncurated ImageNet samples from our model with $P = 4$ and classifier-free guidance scale 1.5. The class labels are: goldfish, arctic fox, monarch butterfly, african elephant, flamingo, tennis ball, cheeseburger, fountain, balloon, tabby cat, lorikeet, agaric.}
\label{imagenet1.5}
\end{measuredfigure}
\end{center}
\end{figure}

\begin{figure}
\begin{center}
\begin{measuredfigure}
\includegraphics[scale=0.3]{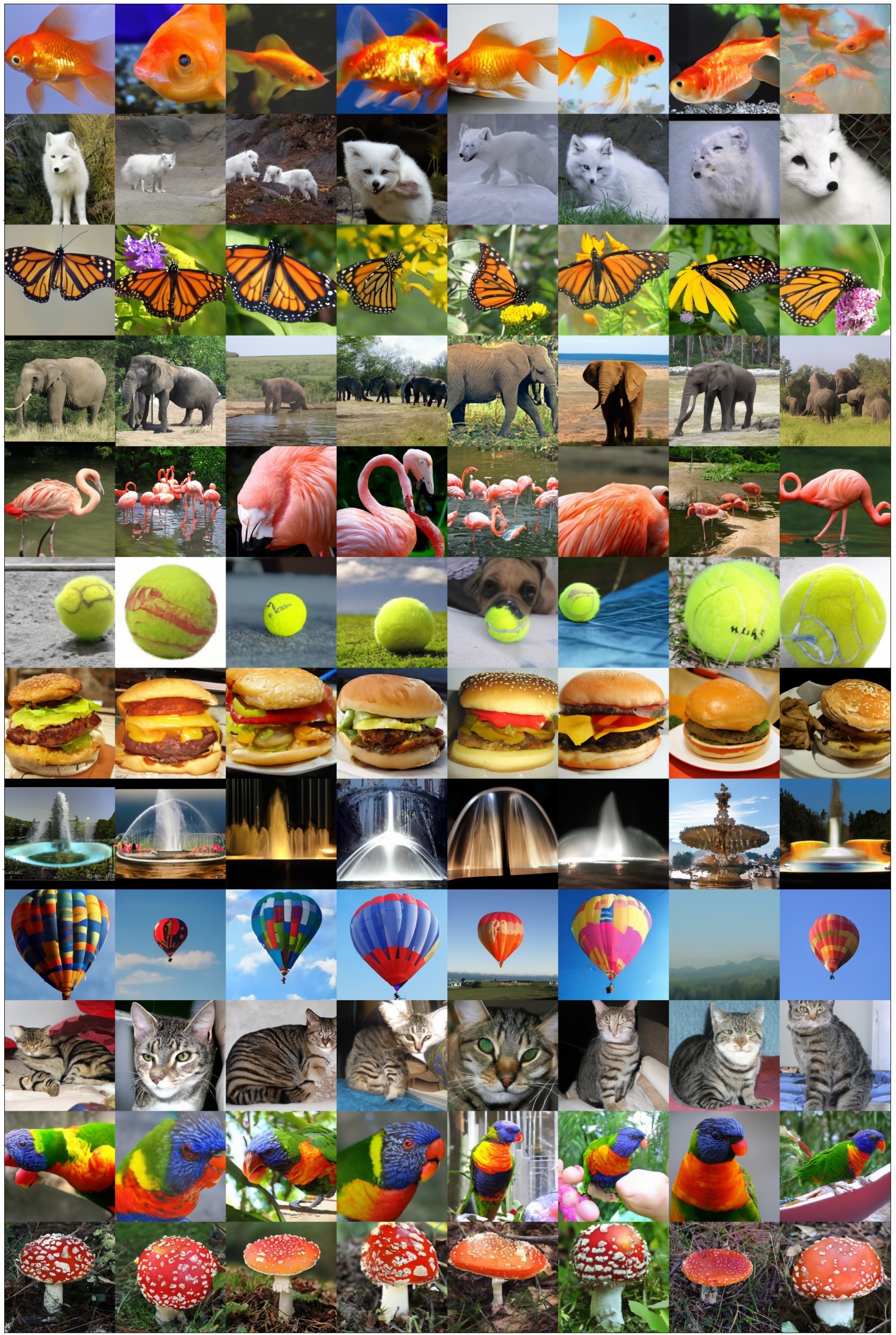}
\vspace*{-0.3cm}
\caption{Uncurated ImageNet samples from our model with $P = 4$ and  classifier-free guidance scale 2.25. The class labels are: goldfish, arctic fox, monarch butterfly, african elephant, flamingo, tennis ball, cheeseburger, fountain, balloon, tabby cat, lorikeet, agaric.}
\label{imagenet2.25}
\end{measuredfigure}
\end{center}
\end{figure}

\begin{figure}
\begin{center}
\begin{measuredfigure}
\includegraphics[scale=0.45]{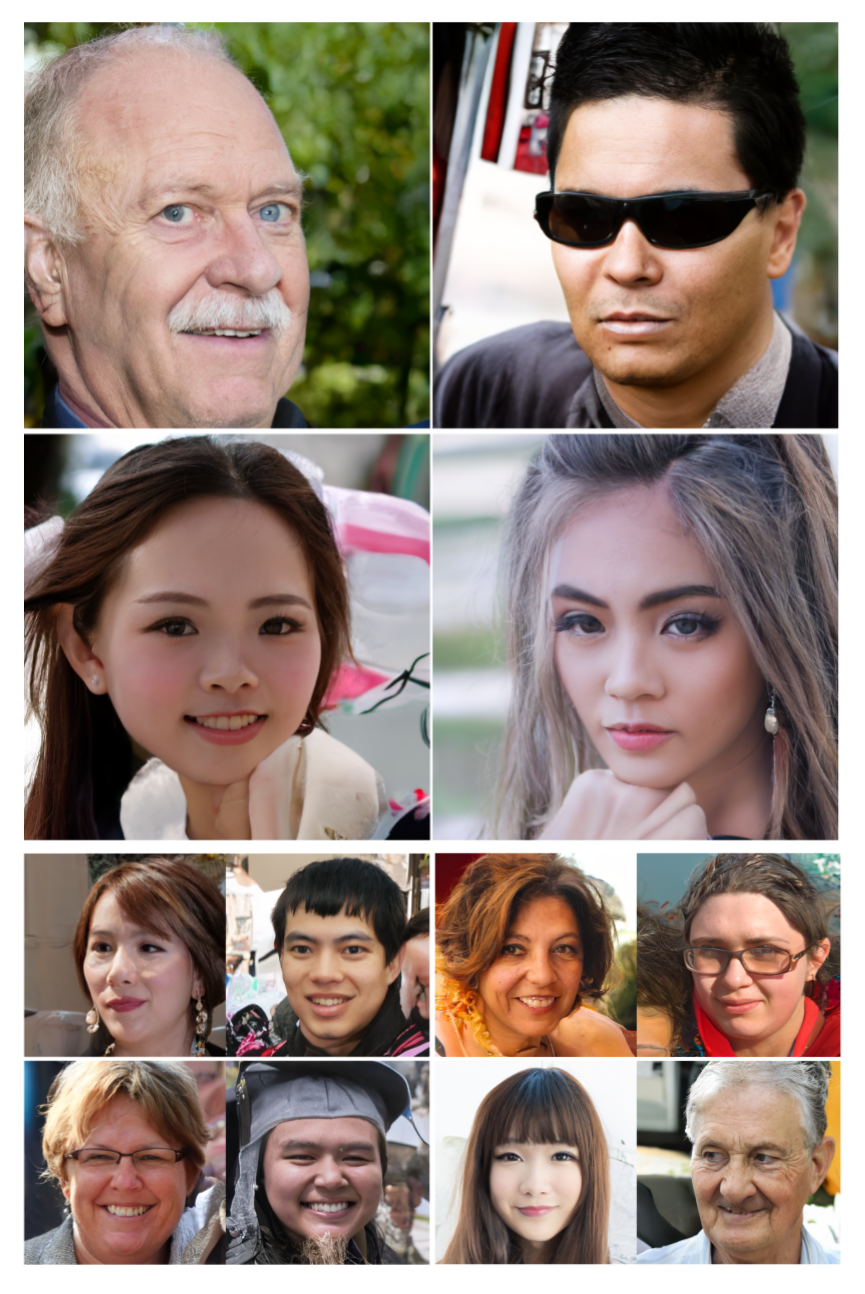}
\vspace*{-0.3cm}
\caption{Uncurated FFHQ samples from our model with $P = 4$. We used 1000 sampling steps for all images.}
\label{ffhqfig}
\end{measuredfigure}
\end{center}
\end{figure}

\end{document}